\documentclass{egpubl}
\usepackage{pg2026}

\SpecialIssuePaper         

\CGFccby

\usepackage[T1]{fontenc}
\usepackage{dfadobe}  

\usepackage{cite}  
\usepackage{soul}
\usepackage{url}
\usepackage[utf8]{inputenc}
\usepackage[small]{caption}

\usepackage{booktabs}
\usepackage{algorithm}
\usepackage{algorithmic}
\usepackage[switch]{lineno}

\usepackage{multirow}%
\usepackage{amsmath,amssymb,amsfonts}%
\usepackage{mathrsfs}%
\usepackage{enumitem}
\usepackage{tabularx}
\usepackage{array}

\usepackage{xcolor}
\definecolor{mygreen}{RGB}{204, 255, 229}
\definecolor{myblue}{HTML}{5364cc}

\BibtexOrBiblatex
\electronicVersion
\PrintedOrElectronic
\ifpdf \usepackage[pdftex]{graphicx} \pdfcompresslevel=9
\else \usepackage[dvips]{graphicx} \fi

\usepackage{egweblnk} 
\title[Q-MAGIC]%
      {Q-MAGIC: Quad-only Mesh Autoregressive Generation via Singularity-Aware Circular Relative Offset Tokenization}

\author[Yuming Zhao et al.]
{
\parbox{\textwidth}{\centering
Yuming Zhao$^{1}$\thanks{Equal contribution.},
Zezeng Li$^{2,\dagger}$,
Weimin Wang$^{1}$\thanks{Corresponding authors.},
Xiang Wu$^{1}$,
Na Lei$^{1,\ddagger}$
}
\\
\parbox{\textwidth}{\centering
$^1$Dalian University of Technology, China\\
$^2$École Centrale de Lyon, France
}
}

\begin{document}


\maketitle

\begin{abstract}
Autoregressive models have proven effective for triangular mesh generation by transforming it into predicting the next face, converting the non-differentiable connectivity prediction into a sequence of token predictions. However, generating quad-only meshes remains challenging because quad faces impose stricter topological constraints than triangles. The model must not only serialize quad faces into a valid token sequence, but also support unambiguous connectivity reconstruction during de-serialization. To address this, we propose \textbf{Q-MAGIC}, a novel autoregressive model for quad mesh generation.
We first introduce a tailored serialization scheme for quad meshes, which requires at least two new vertices for the next face and is less flexible than triangles due to strict wiring constraints. To meet these requirements and reduce serialization length, we propose a singularity-aware serialization that prioritizes wiring from singular points, leading to more coherent traversals. 
Additionally, we propose a circular relative offset tokenization that encodes geometry mainly via local relative offsets.
Built upon this representation, an autoregressive Transformer learns the underlying distribution of quad meshes via next-token prediction, enabling effective quad mesh generation.
Extensive experiments demonstrate that Q-MAGIC produces more compact sequences while consistently improving geometric fidelity and mesh quality compared to state-of-the-art methods.

\ccsdesc[500]{Computing methodologies~Shape modeling}
\ccsdesc[500]{Computing methodologies~Computer vision}
\ccsdesc[500]{Computing methodologies~Neural networks}

\printccsdesc   
\end{abstract}  
\section{Introduction}


Quad meshes are a widely used surface representation in production workflows such as animation, CAD, and physical simulation, where pure quadrilateral meshes are particularly favored for their compatibility with standard modeling pipelines and support for downstream applications, including computational mechanics and geometric analysis. Consequently, developing robust and practical methods for generating high-quality pure quad meshes remains an important problem in both academia and industry.

Most current methods generate quad meshes via a two-stage pipeline. These methods first create a triangle mesh and then apply quadrangulation or remeshing as post-processing. This intermediate triangle mesh can be produced either from continuous 3D representations~\cite{NeRFMeshing,NVDiffRec,DeepSDF,neus,Hong20243DTopiaLT,xu2024instantmeshefficient3dmesh} or from triangle-based autoregressive mesh generators~\cite{bpt,MeshAnythingV2}. However, this conversion adds extra stages and heuristics, and it does not reliably guarantee quad-only outputs, often requiring additional cleanup or refinement. 

Inspired by triangular mesh generation, Liu et al.~\cite{QuadGPT} proposed a mixed triangle and quad face mesh generative model via a united hybrid polygon mesh serialization approach. While this serialization method is capable of handling both triangular and quadrilateral faces simultaneously, it suffers from two key limitations. First, it fails to account for the unique mesh wiring constraints and singular structures inherent to quadrilaterals. Second, the excessively redundant sequence length substantially increases the difficulty of model training and optimization.

Compared to triangular meshes, quad meshes are more imbalanced in structure. Most faces consist of regular valence-4 connectivity, yet the global layout is controlled by a small set of singular vertices and their neighborhoods. These singular structures steer the overall edge flow and indicate where regularity must break. Learning a quad mesh as a uniform sequence over-emphasizes the regular patterns and underfits these rare singularities. As a result, errors in early singularity decisions can propagate along loops, producing inconsistent layouts and reduced editability.

To address these issues, we propose Q-MAGIC, a quad-only autoregressive model that follows a layout-first design. We partition a mesh into patches and serialize them in a singularity-aware order, so topology-critical regions are generated early. Within each patch, our method predicts geometry using circular relative offsets. This representation reduces dynamic range and redundancy, improves robustness to local mesh-density variations, leading to more stable training and higher-quality quad generation. In summary, our contributions are as follows:
\begin{itemize}[leftmargin=*, itemsep=0pt, topsep=2pt, parsep=0pt, partopsep=0pt]
    \item We propose Q-MAGIC, to our knowledge, the first autoregressive Transformer framework for quad-only mesh generation. 
    \item We propose a singularity-aware traversal to prioritize topology-critical wiring, shorten serialization sequences, and improve mesh validity.
    \item We introduce a circular relative offset tokenization to encode geometry and handle mesh-density variation robustly.
    \item Extensive experiments show that Q-MAGIC improves mesh validity and reconstruction quality over state-of-the-art methods. 
\end{itemize}
\section{Related Works}

\subsection{Autoregressive Model for Triangle or Mixed Mesh}  

Following MeshGPT~\cite{MeshGPT}, recent autoregressive mesh generation methods have moved towards stronger conditional controllability and better scalability. Some methods introduce richer shape conditions or control signals to impose stronger constraints on the output without changing the autoregressive paradigm ~\cite{meshanything,MeshTron}.  Other methods address the bottleneck of training and inference caused by long sequences, proposing scalable network architecture and training designs for meshes with more patches and higher resolution~\cite{MeshTron,MeshXL,pivotmesh}.

Another important direction is mesh serialization, which directly affects sequence length, local connectivity modeling, and generation stability. EdgeRunner~\cite{edgerunner} borrows the idea of connectivity compression from EdgeBreaker~\cite{edgebreaker} to achieve lossless patch compression and reduce long-range dependencies. MeshAnythingV2~\cite{MeshAnythingV2} uses adjacent-face encoding to reduce vertex redundancy. OCTGPT~\cite{OCTGPT} uses a multi-scale structure for scalable 3D shape serialization. BPT~\cite{bpt} compresses meshes at both vertex and face levels via blocked and patchified representations to reduce sequence length. MeshRipple~\cite{MeshRipple} aligns the generation order with surface topology through frontier-aware traversal, and Mesh Silksong~\cite{Mesh-SilkSong} emphasizes topology-preserved serialization. FACE~\cite{FACE} represents each face as a compact token, while SATO~\cite{SATO} introduces strip-based tokenization for unified triangle and quadrilateral representation.

Deepmesh and Mesh-RFT~\cite{Deepmesh,meshrft} further explore preference-based post-training (e.g., DPO~\cite{DPO}) to improve the generation quality. Researchers are also exploring accelerated decoding and progressive level-of-detail generation to achieve a better balance between inference speed and output quality ~\cite{flashmesh,Mesh-RAG,xspechmesh,Zhang_2025_ICCV_VertexRegen}.

Recently, autoregressive modeling has been extended to quadrilateral meshes. QuadGPT~\cite{QuadGPT} and Mesh-Pro\cite{Mesh-Pro} present end-to-end quadrilateral dominant mesh generation frameworks that employ unified hybrid polygon serialization and reward-based refinement strategies. QuadLink~\cite{QUADLINK} formulates quad-dominant mesh generation as a point-relation learning problem, predicting vertices and face centroids and assembling polygonal faces through centroid-conditioned vertex links and a quad-first strategy. In general, existing autoregressive methods are either dominated by triangular meshes or produce hybrid meshes, whereas native quad-only mesh generation remains underexplored.

\subsection{Quad Mesh Generation}

Quad mesh generation has long been studied in computer graphics, with traditional methods mainly formulated as remeshing, layout extraction or field-aligned optimization. For example, Marinov and Kobbelt~\cite{Marinov2004DirectAQ} proposed anisotropic quad-dominant remeshing, while Pietroni et al.~\cite{pietroni2021reliable} extracted field-coherent quad layouts from cross fields.
Although these methods can produce high-quality quad structures, they typically rely on carefully constructed inputs, such as clean surfaces, reliable directional fields, or well-initialized layouts. Moreover, since quad connectivity is mainly obtained through hand-designed remeshing or layout extraction procedures, these methods are less flexible in modeling diverse quad-only meshes in a data-driven generative setting.

Beyond traditional optimization-based pipelines, recent learning-based methods have also explored quad mesh generation, although most of them still rely on intermediate fields, templates, or local assembly procedures. Many works first predict a local directional frame field for quadrangulation and then generate quad meshes via field-aligned parameterization or subsequent remeshing. This pipeline imposes strong requirements on the predicted field, which should be globally smooth for stable parameterization, while still aligning with geometric features to support reliable quad extraction~\cite{yu2025neuralpolyvector,NeurCross,crossgen,NeuFrameQ,qi2025rffmeshing, TopGen,crosslift}. 
Another line of work builds on PDE-constrained formulations, e.g., incorporating PDE residuals into the learning objective to learn mappings from a parametric domain to the physical domain for structured mesh generation~\cite{MGNet,3DMeshNet}. MeshONet~\cite{MeshONet} further formulates structured mesh generation as an operator-learning problem for mapping computational domains to physical domains. Other approaches use flow or ODE-based template deformations to fit geometry while largely preserving topology~\cite{Neural_Mesh_Flow,CorticalFlow,zhang2024physics}. Meanwhile, some approaches cast meshing as a sequential decision-making problem and use reinforcement learning to optimize element placement or advancing-front updates, with rewards designed around element quality and topological preferences~\cite{Reinforcement-learning,SRL-Assisted-AFM,Lin2020Modeling3S}. Overall, these methods tend to work best in simple planar geometries or topologies with clean inputs, and their applicability remains limited for complex shapes.

In addition, researchers have explored directly generating quad meshes from point clouds. Point2Quad~\cite{point2quad} directly produces quad-only surface meshes by predicting and selecting local quadrilateral faces for mesh assembly. 
However, it follows a candidate-face prediction paradigm. Therefore, the recoverable quad connectivity is bounded by whether valid faces are included in this candidate set. Our method instead represents quad meshes as autoregressive sequences, where quad faces are generated through the learned sequence and the corresponding de-serialization rule, reducing the reliance on pre-enumerated local face candidates.
\begin{figure*}[t]
    \centering
    \includegraphics[width=.9\textwidth]{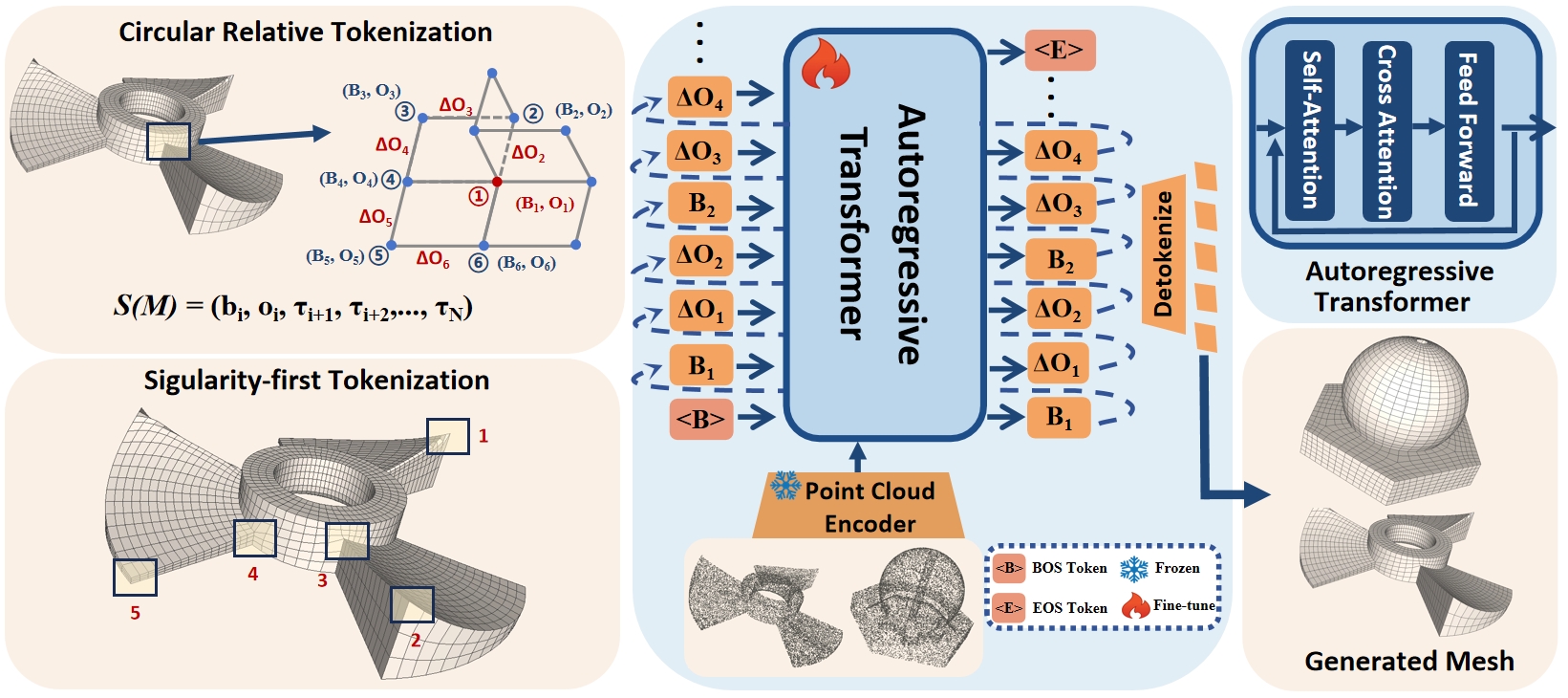}
    \captionsetup{skip=2.5pt}
    \caption{Overview of the Q-MAGIC. We adopt a singularity-aware serialization strategy that prioritizes critical singularities in the sequence. In our token representation, geometric information is mainly encoded by circular relative offsets, and block tokens are inserted only when necessary to re-anchor the spatial location, enabling a compact representation. During inference, Q-MAGIC conditions on point clouds and autoregressively generates quad meshes.}
    \label{fig:pipeline}
\vspace{-5mm}
\end{figure*}

\section{Method}\label{sec3}

As shown in Fig.~\ref{fig:pipeline}, the proposed Q-MAGIC centers on a compact serialization for quad-only meshes, which transforms mesh geometry and connectivity into a token sequence for autoregressive generation. 
While Q-MAGIC adopts the block--offset indexing introduced by BPT\cite{bpt}, BPT is designed for triangular meshes, whose local serialization exploits the fact that every pair of vertices within a triangle is directly connected by an edge. This property does not hold for quadrilateral faces, where diagonally opposite vertices are not edge-connected. Consequently, directly extending BPT-style traversal to quad meshes becomes insufficient for preserving valid connectivity, particularly around singular regions with non-consecutive local face configurations.
The serialization follows a singularity-aware traversal and uses circular relative offset tokens within local patches to produce shorter sequences while retaining topology-critical structures.

\subsection{Naive Quad Mesh Serialization}

We formulate the quad mesh generation task as a sequence prediction problem, transforming a mesh $\mathcal{M}$ to a one-dimensional sequence of discrete tokens. 
Following triangular mesh serialization, a naive quad mesh serialization represents a mesh as an ordered collection of faces together with their vertex coordinates.
Let $\mathcal{M}=\{f_i\}_{i=1}^N$ denote a mesh with $N$ quadrilateral faces, where the $i$-th face is $f_i=(v_{i1},v_{i2},v_{i3},v_{i4})$ and each vertex $v_{ij}=(x_{ij},y_{ij},z_{ij})\in\mathbb{R}^3$.
To make the continuous geometry amenable to autoregressive modeling, all vertex coordinates are first rescaled to the fixed range $[-1,1]$, denoted by $\hat{v}_{ij}=(\hat{x}_{ij},\hat{y}_{ij},\hat{z}_{ij})$,
and each axis is uniformly quantized with resolution $r$. Specifically, the quantized coordinate is computed by
\begin{equation}
\label{eq:coord_quant}
\tilde{x}_{ij}=\left\lfloor \frac{\hat{x}_{ij}+1}{2}(r-1)+\frac{1}{2}\right\rfloor \in \{0,1,\dots,r-1\},
\end{equation}
and the same operation is applied to $\hat{y}_{ij}$ and $\hat{z}_{ij}$. This yields quantized vertices $\tilde{v}_{ij} = (\tilde{x}_{ij}, \tilde{y}_{ij}, \tilde{z}_{ij})$.

All integer vertices are then sorted in the lexicographic order of $Z$, $Y$, $X$, and assigned global indices accordingly, and quadrilateral faces are sorted by their vertex indices to obtain a globally ordered face sequence $F = (f_1, f_2, \dots, f_N)$.

Based on this ordering, faces are traversed in sequence. Within each face, the four vertices are visited in a fixed order. The three quantized coordinates of every vertex are listed consecutively, resulting in a length-$12N$ discrete token sequence
\[    S(\mathcal{M}) = (\tilde{x}_{11}, \tilde{y}_{11}, \tilde{z}_{11},\;
            \tilde{x}_{12}, \tilde{y}_{12}, \tilde{z}_{12},\;
            \dots,\;
            \tilde{x}_{N4}, \tilde{y}_{N4}, \tilde{z}_{N4}).
            \]
For a quadrilateral mesh with $N$ faces, this naive scheme yields a sequence of length $L_{\text{naive}} = 12N$ over a vocabulary
size of $r$. This highly redundant and lengthy sequence substantially elevates the complexity of model learning tasks.


\begin{figure*}[t]
    \centering
    \includegraphics[width=0.9\textwidth]{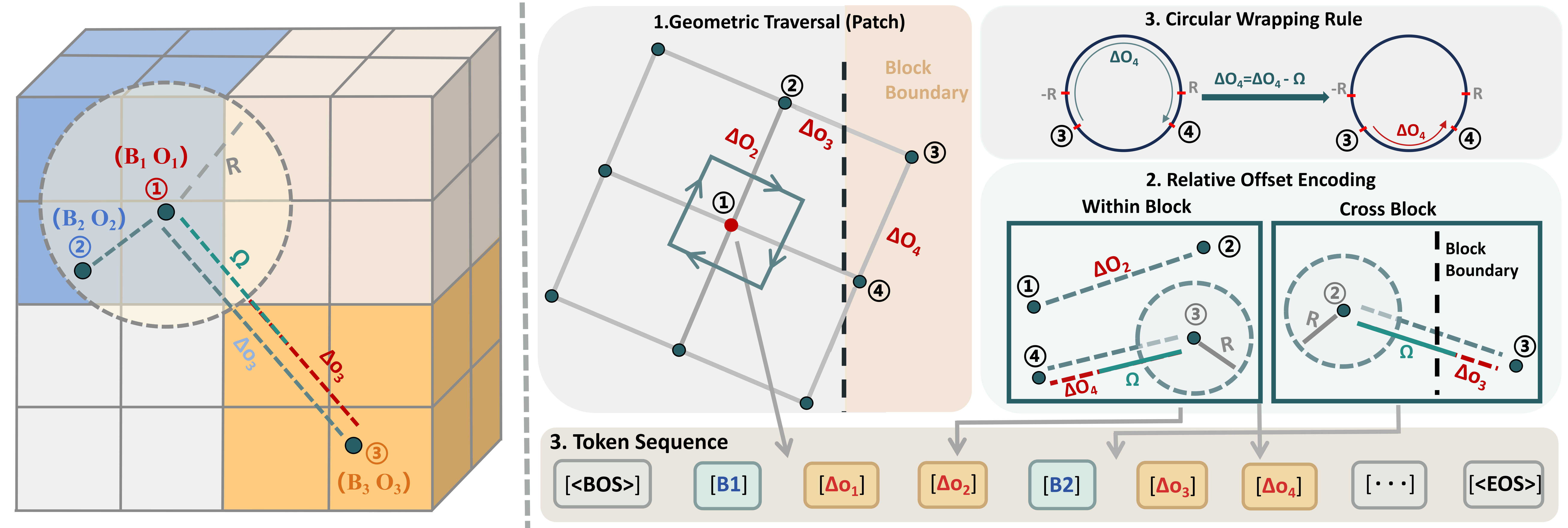}
    \captionsetup{skip=0.05pt}  
    \caption{Overview of circular relative offset tokenization. Left: each vertex is represented by a block–offset pair $(B, O)$. Right: For intra-block moves, we emit only $\Delta o_n$ (wrapping to the bounded interval when needed); when crossing a block boundary, we emit the new block index $B$ and then continue with relative offsets. Bottom: the resulting token sequence with BOS/EOS, consisting of block tokens and circular relative offset tokens. }
    \label{fig:tokenizer}
    \vspace{-5mm}
\end{figure*}

\subsection{Singularity-aware Serialization}
\label{subsec:serialization}

\textbf{Block--offset indexing.}
Although the naive quad mesh serialization described above is simple and directly compatible with autoregressive models, it treats each quantized coordinate as an independent token and repeats shared vertices for every incident face. Existing methods, such as QuadGPT~\cite{QuadGPT}, adopt similar coordinate-based tokenizations, which can result in long and redundant sequences and a vocabulary size that grows cubically with the coordinate resolution. To address these issues, we follow BPT\cite{bpt} and convert quantized coordinates into a block–offset
representation, as shown in Fig.~\ref{fig:tokenizer}.

Given the vertex sequence $S$, we discretize each 3D position into a block-offset index.
Along each coordinate axis, we partition the coordinate range into $B$ blocks and subdivide each block into $O$ offsets. So the resolution per axis is $B\cdot O$.
For a quantized vertex $\tilde{v}_{ij} = (\tilde{x}_{ij}, \tilde{y}_{ij}, \tilde{z}_{ij})$
with each integer coordinate in
$\{0,\dots,B\cdot O-1\}$, the block-wise indexing $(b_i, o_i)$ is computed as
\begin{equation}
\label{eq:block_offset}
\begin{aligned}
    b_i &= (\tilde{x}_i \,\texttt{|}\, O)\, \cdot B^2 + (\tilde{y}_i \,\texttt{|}\, O)\, \cdot B + (\tilde{z}_i \,\texttt{|}\, O),\\
    o_i &= (\tilde{x}_i \% O)\, \cdot O^2 + (\tilde{y}_i \% O)\, \cdot O + (\tilde{z}_i \% O),
\end{aligned}
\end{equation}

\noindent where \texttt{|} denotes integer division and \% denotes the modulo operation.
Here $b_i \in [0, B^3 - 1]$ is the block index and $o_i \in [0, O^3 - 1]$ is the absolute
offset inside that block. Together, the block-offset pair
$(b_i, o_i)$ uniquely encodes a quantized 3D coordinate with per-axis
resolution $r = B \cdot O$. The overall tokenization process is summarized in Algorithm~\ref{alg:quad_bpt}.
\begin{algorithm}[!tbp]
\caption{Singularity-aware Quad Mesh Tokenization}
\label{alg:quad_bpt}
\small
\begin{algorithmic}[1]
\STATE \textbf{procedure} Tokenization$(\mathcal{M})$
\STATE $\mathcal{P}\gets\emptyset$ \hfill $\triangleright$ List of patches
\WHILE{$\mathcal{M}\neq\emptyset$}
    \STATE $v_c\gets\textsc{SelectCenter}               (\mathcal{V}_{\mathrm{irr}}\succ\mathcal{V}_{\mathrm{reg}},\mathcal{M})$
    \STATE $P_c\gets\textsc{Aggregate}(v_c)$
    \STATE $P_c\gets\textsc{Blockwise-Index}(P_c)$ \hfill $\triangleright$ Eq.~\ref{eq:block_offset}
    \STATE $P_c\gets\textsc{Relative-Offset}(P_c)$ \hfill $\triangleright$ Eq.~\ref{calculate_delta_o}
    \STATE $P_c\gets\textsc{Dual-Block}(P_c)$
    \STATE $\mathcal{P}\gets\mathcal{P}\cup P_c,\quad \mathcal{M}\gets\mathcal{M}-P_c$
\ENDWHILE
\STATE \textbf{return} $\mathcal{P}$
\end{algorithmic}
\end{algorithm}

\textbf{Singularity-aware patch construction.}
Local topological complexity can be characterized by the valence of a vertex. We refer to vertices whose valence is not equal to 4 as singular vertices, and denote the set of such vertices as $V_{\mathrm{irr}}$.
During traversal, patch centers are first selected from $V_{\mathrm{irr}}$ until all singular vertices are covered.

\begin{figure}[htbp]
    \centering
    \includegraphics[width=0.98\linewidth]{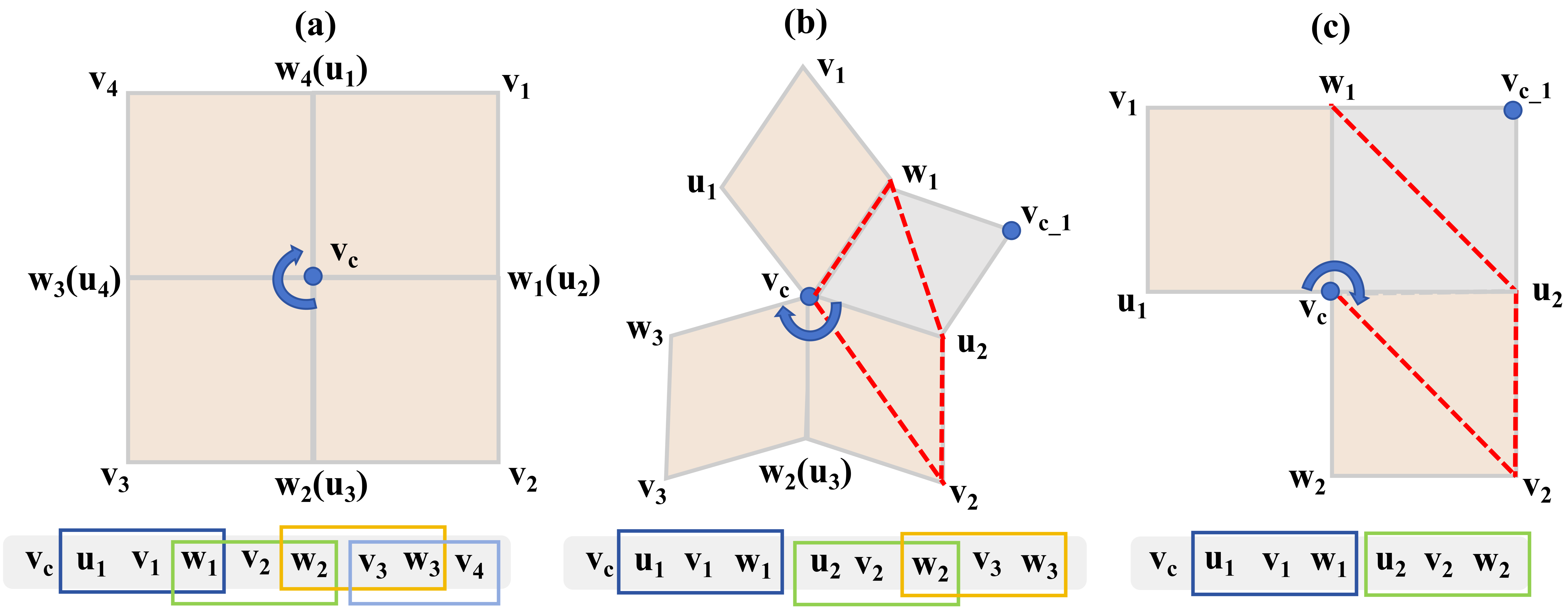}
    \captionsetup{skip=2.5pt}
    \caption{De-serialization in Q-MAGIC. (a) Consecutive records recover the shared vertex from the last vertex of the previous face. (b,c) Directly connecting adjacent tokens for non-consecutive records may generate invalid red dashed quads. 
    We resolve this issue via a novel de-serialization scheme. Quads are reconstructed from the recorded non-center vertices under the face of the current local patch, ensuring consistent local topology.}
    \label{fig:cases}
    \vspace{-5mm}
\end{figure}
After the center vertex $c$ is determined, we collect all quadrilateral faces incident to $c$ and aggregate them into a local patch $\mathcal{P}(c)$. Each face in the patch is written as
\[
    f_i=(c,u_i,v_i,w_i)
\]
where $u_i$ and $w_i$ are the two vertices connected to $c$ by real mesh edges, and $v_i$ is the vertex diagonal to $c$. In the traversal sequence of $\mathcal{P}(c)$, the vertex immediately following the center $c$, which we call the starting vertex, is selected only from $\{u_i\}\cup\{w_i\}$.

Fig.~\ref{fig:cases} illustrates three representative local quad-face layouts and their corresponding serialized vertex sequences. Case (a) shows a regular four-valent configuration, where the incident quad faces form a consecutive layout around the center vertex. In this case, the BPT-style local traversal can be naturally extended to quad meshes by recording the non-center vertices in order.

However, cases (b) and (c) show two typical singular configurations with non-consecutive local layouts. The gray faces indicate quad faces that have been assigned to other center vertices and should not be serialized under the current center vertex. After excluding these faces, the remaining incident faces around the current center vertex no longer form a simple consecutive sequence. In particular, the last vertex of the previous face may not coincide with the first vertex of the next face. Therefore, directly applying the triangle-based BPT serialization rule to quad meshes is insufficient.

To handle such cases, we first identify all vertices that share an edge with the center vertex in the current local patch. Among these adjacent vertices, we select the first vertex encountered along the specified traversal direction as the starting vertex.
From this vertex, we traverse the valid quad faces around the center vertex clockwise. For consecutive faces, if the first non-center vertex of the current face matches the last non-center vertex of the previous face, only the remaining two non-center vertices are appended; otherwise, a new local record is started with all three non-center vertices of the current face.

Similarly, when deserializing consecutive records, the shared vertex is recovered from the last vertex of the previous face. For a non-consecutive record, a new quad face should be initialized from its three recorded non-center vertices. If one directly follows the original BPT decoding rule and simply connects locally adjacent tokens, the decoder may incorrectly reconstruct the red dashed quads shown in Fig.~\ref{fig:cases}~(b) and (c). These quads should not exist under the current center vertex. Therefore, our quad-aware de-serialization follows the inverse of the serialization rule while respecting the face ownership of the current local patch.
The corresponding de-tokenization procedure is summarized in Algorithm~\ref{alg:quad_detoken}.
\begin{algorithm}[!tbp]
\caption{Quad Mesh De-tokenization}
\label{alg:quad_detoken}
\small
\begin{algorithmic}[1]
\STATE \textbf{procedure} De-serialization$(\mathbf{S})$
\STATE $\mathcal{M}\gets\emptyset$
\STATE $\{S_c\}\gets\textsc{Split}(\mathbf{s},\mathtt{sp})$
\WHILE{$S_c\neq\emptyset$}
    \STATE $(v_c,P_c)\gets\textsc{Parse}(S_c)$
    \STATE $P_c\gets\textsc{InvDual}(P_c)$
    \STATE $P_c\gets\textsc{RecoverO}(P_c)$ \hfill $\triangleright$ Eq.~\ref{eq:offset_recursion}
    \STATE $P_c\gets\textsc{InvIndex}(P_c)$ \hfill $\triangleright$ Eq.~\ref{eq:block_offset}
    \STATE $\mathcal{M}\gets\mathcal{M}\cup\textsc{RebuildQuad}(v_c,P_c)$
\ENDWHILE
\STATE \textbf{return} $\mathcal{M}$
\end{algorithmic}
\end{algorithm}
\vspace{-2mm}

\subsection{Circular Relative Offset Tokenization}
As illustrated in Fig.~\ref{fig:tokenizer}, we encode offsets as relative moves on a discrete circle, mapping raw differences into a bounded interval.

We reparameterize the offset sequence by replacing absolute offsets $\{o_n\}$ with wrapped differences between consecutive offsets.
Let the offset space size be $|\Omega|=O^3$ and define
\[
    \Delta o_n = o_n - o_{n-1}, \qquad R = \bigl\lfloor \Omega / 2 \bigr\rfloor .
\]
The differences $\Delta o_n$ are folded into a symmetric interval:
\begin{equation}
\label{calculate_delta_o}
   \Delta o_n' =
    \begin{cases}
    \Delta o_n - \Omega, & \text{if } \Delta o_n > R,\\[4pt]
    \Delta o_n + \Omega, & \text{if } \Delta o_n < -R,\\[4pt]
    \Delta o_n,     & \text{otherwise}.
    \end{cases}
\end{equation}
ensuring that $\Delta o_n'$ falls within the range $[-R, R]$.
The absolute offsets can be recovered recursively as
\begin{equation}
\label{eq:offset_recursion}
    o_n = o_{n-1} + \Delta o_n' \pmod{\Omega}.
\end{equation}
Equivalently, for any reference index $i \le n$,
\begin{equation}
\label{eq:offset_sum}
    o_n = o_i + \sum_{k=i+1}^{n} \Delta o_k' \pmod{\Omega}.
\end{equation}
This circular relative offset representation has two immediate properties. Firstly, under the relative encoding, each token only needs to specify a local displacement
$\Delta o_n'$ within the bounded interval $[-R, R]$, rather than selecting an absolute index from
the entire offset space $\{0,\dots,\Omega-1\}$. 
Secondly, under this circular relative offset parametrization, the geometry of an intra-block vertex
sequence is mainly determined by the sequence of relative offsets $\Delta o_n'$, and is
independent of the specific value of the initial offset. If the starting offset is
changed, all subsequent vertices are translated consistently, while their relative configuration
remains unchanged

Since consecutive vertices often fall into the same block, we adopt a run-based circular relative offset encoding that keeps an explicit block index only when necessary.
As illustrated in Fig.~\ref{fig:tokenizer}, the encoding covers three typical cases.

\textbf{Intra-block and representable offset.}
When two consecutive vertices $v_n$ and $v_{n-1}$ lie in the same block with $|\Delta o_n|\le R$, we compute $\Delta o_n = o_n - o_{n-1}$ and append it directly as a circular relative offset token, without emitting a block token or applying circular wrapping.
In the \emph{Intra-block} example of Fig.~\ref{fig:tokenizer}~(c), vertices $1$ and $2$ stay in the same block and the displacement falls within the preset radius. Thus, we directly append the circular relative offset token $\Delta o_2$ to the sequence.

\textbf{Intra-block but out-of-range.}
Even when $b_n=b_{n-1}$ holds, the raw intra-block displacement $\Delta o_n=o_n-o_{n-1}$ can still exceed the preset radius $R$. As shown by vertices $3$ and $4$ in Fig.~\ref{fig:tokenizer}~(c), the two vertices lie in the same block, yet their offset difference is larger than $R$. We therefore apply the circular mapping to wrap this difference back into the representable range of radius $R$, obtaining a bounded displacement $\Delta o_4'$, and append the resulting circular relative offset token to the sequence.

\textbf{Cross-block.}
When the traversal crosses a block boundary, an explicit block index is required to re-anchor the location.
In the \emph{Cross Block} example of Fig.~\ref{fig:tokenizer}~(c), vertices $2$ and $3$ fall into different blocks ($b_3\neq b_2$). Therefore, we append the block index $b_3$ together with its bounded offset $\Delta o_3'$ to the sequence, since the block index cannot be omitted across blocks. 

Formally, given the original block--offset representation $(b_i,o_i,\; b_{i+1},o_{i+1},\dots,b_N,o_N)$, we rewrite it as
\begin{equation}
S(\mathcal{M})= (b_i,o_i,\; \tau_{i+1},\; \tau_{i+2},\; \dots,\; \tau_N),
\label{eq:rel_offset_run}
\end{equation}
where each emitted token $\tau_k$ follows a single rule:
\begin{equation}
\tau_k =
\begin{cases}
\Delta o_k', & \text{if } b_k=b_{k-1}\ \\
(b_k,\Delta o_k'),   & \text{otherwise}.
\end{cases}
\label{eq:tau_def}
\end{equation}

As shown in Fig.~\ref{fig:tokenizer}~(e), we encode the fine-grained location using circular relative offset tokens. When the relative offset exceeds the radius $R$, we wrap it back to $[-R, R]$ on the discrete circle. Only when the traversal enters a new block do we insert an explicit block token $b_k$ to re-anchor the coarse location.

To quantify the compactness of our serialization scheme, we calculate the average sequence length produced by different serialization schemes.
We apply the serialization scheme of each method to the same set of meshes and report the resulting mean sequence length. Our serialization produces the shortest token sequence on average among all compared methods,
as shown in Table~\ref{tab:avg_seq_len}.
We further provide an analysis of the sequence composition in the appendix,  which helps explain the observed length differences.
\begin{table}[!tbp]
\captionsetup{skip=2.5pt}
\centering
\footnotesize 
\setlength{\tabcolsep}{1.3mm}
\renewcommand{\arraystretch}{1.15}
\caption{Average sequence length comparison. QGPT, MAV2, EG, and DM denote QuadGPT, MeshAnythingV2, EdgeRunner and DeepMesh. Our serialization yields the lowest token length on the same mesh set.}
\label{tab:avg_seq_len}
\begin{tabular}{c|c|c|c|c|c|c}
\hline
Method & QGPT & MAV2 & EG & DM & BPT & \textbf{Ours} \\
\hline
Avg. Len. & 43106 & 29000 & 28497 & 20675 & 14104 & \textbf{12655} \\
\hline
Comp. Ratio &1 & 0.67 & 0.66 & 0.48 & 0.33
& \textbf{0.29} \\
\hline
\end{tabular}
\vspace{-2mm}
\end{table}

\begin{figure*}[tbp!]
    \centering
    \includegraphics[width=1.0\linewidth]{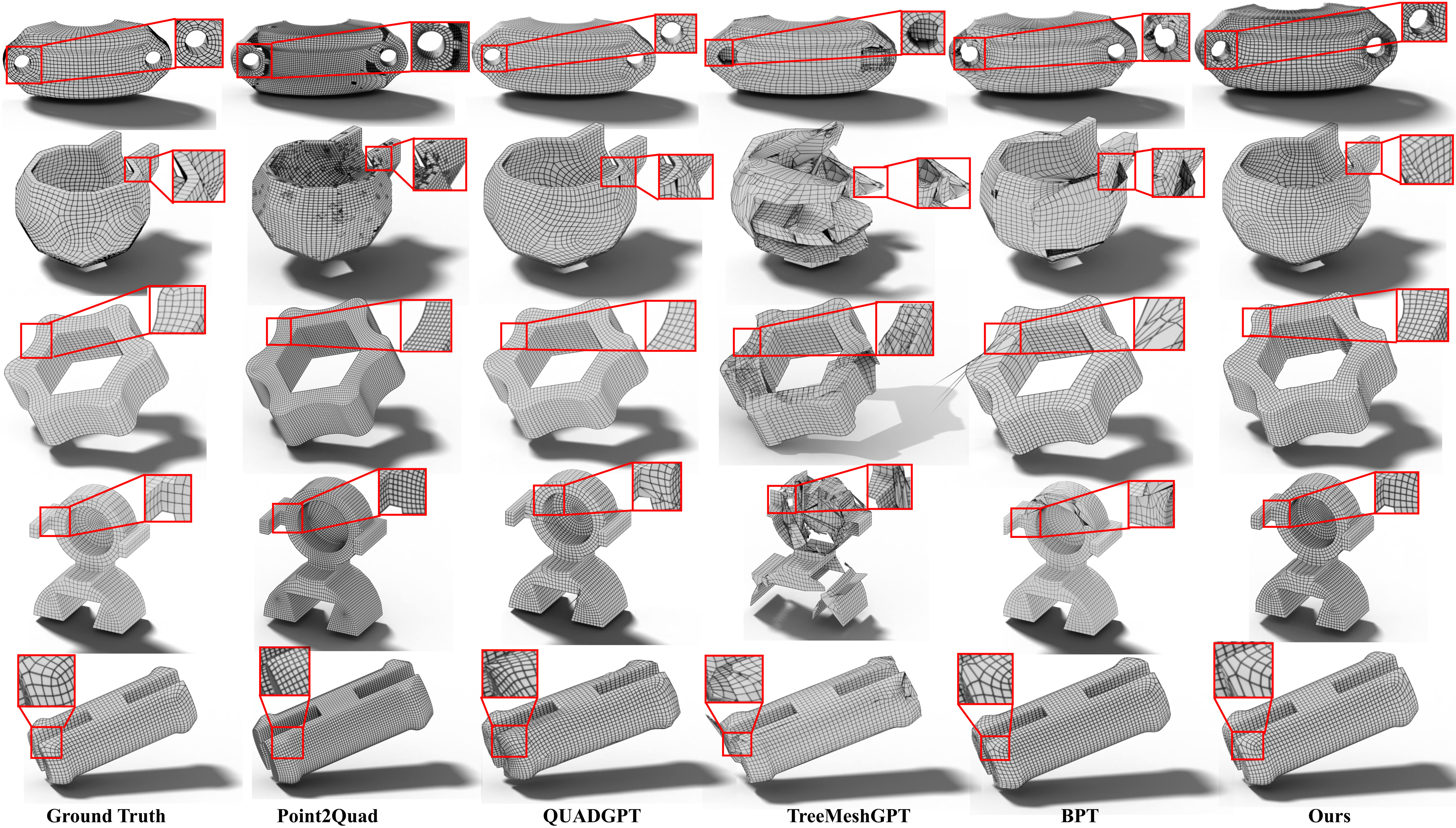}
    \captionsetup{skip=2.5pt}
    \caption{Visualization results and close-up views of generated quad meshes. Our method can better preserve continuous boundary loops and structured quad layouts in detail-sensitive regions, producing more regular and stable quad meshes.}
    \label{fig:visual_compare}
\vspace{-2mm}
\end{figure*}

\section{Experiments}\label{sec4}
\vspace{-1mm}
In this section, we conduct comprehensive experiments to evaluate  Q-MAGIC for quad-mesh generation. Our evaluation emphasizes (i) geometric fidelity and quad-mesh quality and (ii) topological correctness.
We compare  Q-MAGIC against strong baselines, including BPT~\cite{bpt}, TreeMeshGPT~\cite{TreeMeshGPTAM}, QuadGPT~\cite{QuadGPT}, and Point2Quad~\cite{point2quad}, and further validate the contributions of key components via ablation studies.
Specifically, Sec.~\ref{sec:ExperimentSettings} introduces the experimental settings, including the datasets, baselines, metrics, and implementation details. Sec.~\ref{sec:Quantitative} presents the quantitative results for comprehensive comparison, followed by Sec.~\ref{sec:Qualitative}, which provides qualitative results and visual analyses. Finally, we perform ablation studies in Sec.~\ref{sec:Ablation} to assess each core component.

\subsection{Experiment Settings}\label{sec:ExperimentSettings}
\textbf{Datasets.} We construct our experimental dataset with the \textit{obj} files in Thing10K dataset~\cite{pietroni2021reliable} and apply a fixed, automated filtering pipeline to ensure mesh quality and reliable evaluation. We retain only meshes that satisfy the following criteria: all-quadrilateral faces, valid vertex coordinates, no degenerate faces, and no non-manifold edges. Since some meshes have excessively high resolution and an excessive number of faces, we retopologize and simplify them using Quad Remesher~\cite{maxime_quadremesher_2024}, limiting the face count to under 4,700 to guarantee stable training and manageable computational overhead. After filtering and simplification, we obtain 2,100 quad meshes. We randomly split the dataset into 80\% for training and 20\% for testing. To improve robustness to object orientation, we apply rotation augmentation to the training data.

\textbf{Baselines.} We compare  Q-MAGIC with two types of methods. The first type consists of autoregressive triangle or hybrid mesh generators, including BPT~\cite{bpt}, TreeMeshGPT~\cite{TreeMeshGPTAM}, and QuadGPT~\cite{QuadGPT}. For a fair evaluation, we convert the outputs of the autoregressive baselines into quad meshes using the Blender Quad Remesher~\cite{maxime_quadremesher_2024}, an automatic remeshing tool that converts an input surface into a quad-based mesh, and we apply the same post-processing settings across methods. During this quad remeshing process, the algorithm may automatically fill small holes, which can lead to the presence of a limited number of triangular faces in the vicinity of the filled regions. The second type is a direct quad-only reconstruction method, Point2Quad~\cite{point2quad}.  All point cloud conditioned methods take 4,096 uniformly sampled surface points as input. As QuadGPT does not provide a pretrained model, we use the official API (\url{https://3d.hunyuan.tencent.com/}) 
to produce its results. The generation process can be viewed as resembling a topology-optimization-style pipeline.

\textbf{Metrics.}  We assess geometric fidelity using Chamfer distance (CD) and Hausdorff distance~(HD) computed on uniformly sampled 1024 points from the generated and ground-truth meshes, where lower is better. We evaluate topological validity using Watertightness and Manifoldness, where higher is better. We further report the mesh-averaged per-face Scaled Jacobian as a measure of quad element quality, where higher is better.

\textbf{Implementation Details.}
Q-MAGIC is a compact autoregressive Transformer with 8 decoder layers and a hidden size of 1024.
For mesh tokenization, we set the block size to $B=8$ and the offset size to $O=16$, resulting in a per-axis spatial resolution of $B\cdot O=128$. The maximum sequence length is about 20K tokens.

For point-cloud conditioning, we employ a frozen pretrained Michelangelo Encoder to extract 768-dimensional point-cloud features. The extracted features are projected into the 1024-dimensional latent space of the Transformer and incorporated into the autoregressive decoder through cross-attention.

For remeshing, we convert the triangular meshes generated by BPT into quad meshes using Blender Quad Remesher. We set the target face count to 4,500 and the adaptive size to 30\%, allowing the remesher to adjust the local face density according to the geometric characteristics of each mesh.

We first sample 16,384 surface points per shape and randomly subsample 4,096 points during training.
All experiments are trained on a single NVIDIA A100 GPU for approximately 7 days.
The model is trained using the standard next-token cross-entropy loss.
The model is optimized with AdamW using $\beta_1=0.9$ and $\beta_2=0.95$, with a base learning rate of $5\times10^{-5}$ and a batch size of 4.
We adopt a LambdaLR schedule with a linear warmup for the first 10\% of training steps, followed by a cosine decay to a minimum learning rate ratio of 0.5.
During inference, we autoregressively generate mesh tokens using multinomial sampling with top-$k$ sampling ($k=50$) and a temperature of 0.7.

\begin{table}[!tbp]
\captionsetup{skip=2.5pt}
\caption{Quantitative comparison. $\uparrow^a_{b}$ and $\downarrow^a_{b}$ denote that larger and smaller values are better, respectively. Here, $a$ and $b$ represent the upper and lower bounds.SJ, WT, MF, CD, HD, SR and US denote Scaled Jacobian, Watertightness, Manifoldness, Chamfer distance, Hausdorff distance, singularity rate and user study.}
\label{tab:quantitative}
\centering
\setlength{\tabcolsep}{1.0mm}
\renewcommand{\arraystretch}{1.2} 
{\fontsize{8pt}{9.5pt}\selectfont
\begin{tabularx}{0.98\columnwidth}{@{}l|
>{\centering\arraybackslash}X|
>{\centering\arraybackslash}X|
>{\centering\arraybackslash}X|
>{\centering\arraybackslash}X|
>{\centering\arraybackslash}X@{}}
\hline
Model & Point2Quad & TMGPT & QuadGPT & BPT & \textbf{Ours}\\
\hline
SJ$\uparrow^{1}_{-1}$ & 0.9726 & 0.8879 & 0.9727 & 0.9762 & \textbf{0.9829} \\
WT$\uparrow^{1}_{0}$ & 0.9995 & 0.8921 & 0.9836 & 0.9873 & \textbf{0.9996} \\
MF$\uparrow^{1}_{0}$ & \textbf{1} & 0.9989 & 0.9998 & 0.9998 & 0.9999 \\
CD$\downarrow^{\infty}_{0}$  & 0.0717 & 0.1704 & 0.0854 & 0.0719 & \textbf{0.0684} \\
HD$\downarrow^{\infty}_{0}$  & 0.1069 & 0.3255 & 0.1374 & 0.1299 & \textbf{0.1058} \\
SR$\downarrow^{1}_{0}$ & 0.3050  & 0.8715  & 0.2164  & 0.4000   &\textbf{0.2310} \\
US$\uparrow^{10}_{0}$        & 6.95 & 5.88 & 7.48 & 7.88 & \textbf{8.84} \\
\hline
\end{tabularx}
}
\vspace{-3mm}
\end{table}

\subsection{Quantitative Results}\label{sec:Quantitative}
Table~\ref{tab:quantitative} reports the quantitative comparison between Q-MAGIC and Point2Quad, TreeMeshGPT, QuadGPT, and BPT. Overall, Q-MAGIC achieves the best results in geometric fidelity, as measured by CD and HD, and element quality, as reflected by the Scaled Jacobian, while preserving high mesh validity in terms of Watertightness and Manifoldness. These results demonstrate that Q-MAGIC produces more regular quad meshes with more accurate surface fitting, without sacrificing closedness or manifold consistency.

\textbf{Geometric Fidelity}.
We evaluate geometric fidelity using CD and HD, where lower values are better. Q-MAGIC achieves the lowest errors on both metrics, showing that direct quad mesh generation preserves geometric accuracy. Although some baselines obtain competitive results by first generating triangle meshes and then remeshing them into quads, their final quality remains limited by the intermediate triangle representation. Errors in the generated triangle mesh can propagate through remeshing, as observed in TreeMeshGPT. By directly predicting quad structures, Q-MAGIC avoids this propagation and improves geometric fidelity.

\textbf{Mesh Validity}.
Watertightness and manifoldness are used to assess the topological validity of generated meshes, where higher values indicate more closed and manifold structures. As shown in Table~\ref{tab:quantitative}, Q-MAGIC achieves near-perfect scores on both metrics, confirming that direct quad generation preserves strong structural validity. Compared with Point2Quad, Q-MAGIC obtains better geometric fidelity and element quality while maintaining comparable watertightness and manifoldness. This suggests that Q-MAGIC achieves a more favorable balance between surface accuracy, mesh regularity, and topological consistency.

\textbf{Element Quality}.
We evaluate element quality using the Scaled Jacobian, where higher values indicate better-shaped elements. This metric reflects the distortion and regularity of quad faces. Q-MAGIC achieves the highest score, suggesting that its generated quads are closer to ideal elements with lower distortion. This improvement is important for downstream applications such as simulation and geometric processing, since element quality directly affects both practical usability and numerical stability.

\textbf{User Study}.
We conducted a user study to evaluate perceptual mesh quality. For each test case, the outputs of all compared methods were presented in a randomized order with method names hidden, forming a blind evaluation protocol. Testers rated the outputs of all methods on a 0--10 scale, where higher scores indicate better visual quality. Considering the cost of participant involvement, we performed the study on a limited set of ten cases with 50 testers. Although the number of cases is limited, the observed preferences provide supporting evidence for the effectiveness of Q-MAGIC. As shown in Table~\ref{tab:quantitative}, our method achieves the highest mean score, indicating a perceptual preference for our results. This is consistent with the qualitative comparisons, where Q-MAGIC produces more continuous boundary loops, fewer unnecessary openings and holes, and more regular quad layouts with coherent local structures in detail-sensitive regions. These advantages lead to higher subjective ratings.

\subsection{Qualitative Results}\label{sec:Qualitative}
Fig.~\ref{fig:visual_compare} compares Q-MAGIC with Point2Quad, BPT, TreeMeshGPT, and QuadGPT, with the ground-truth quad meshes shown in the first column. For each example, we zoom in on representative regions to examine mesh structure differences at a fine scale. Near boundary openings, Q-MAGIC consistently forms continuous boundary loops and stable quad layouts, with smoother variations in element size and orientation. In contrast, some baselines exhibit broken edge-loop continuity, locally distorted elements, or uneven layouts in the same regions. On large curved surfaces and transition areas, Q-MAGIC produces a more coherent quad flow, with reduced local shear distortion and elements closer to ideal shapes. In the enlarged region of the fourth row, our method avoids unnecessary boundary singularities and yields a more regular local valence pattern and connectivity structure. Overall, our method preserves both geometric adherence and structured quad topology in detail-sensitive regions, resulting in more regular and stable quad meshes.

\begin{table}[!tbp]
\captionsetup{skip=2.5pt}
\caption{Ablation study on patch serialization policies under \textit{circular relative offset} tokenization (Random vs.\ singularity-aware).}
\label{tab:ablation_traversal}
\centering
\footnotesize
\renewcommand{\arraystretch}{1.2}
\begin{tabularx}{0.98\columnwidth}{l|>{\centering\arraybackslash}X|>{\centering\arraybackslash}X}
\hline
Metrics & Random & Singularity-aware \\
\hline
Scaled Jacobian $\uparrow^1_{-1}$                    & 0.8862 & 0.9829 \\
Watertightness $\uparrow^1_{0}$                    & 0.9429 & 0.9996 \\
Manifoldness $\uparrow^1_{0}$                      & 0.9986 & 0.9999 \\
Chamfer distance $\downarrow^{\infty}_{0}$                & 0.2202 & 0.0684 \\
Hausdorff distance $\downarrow^{\infty}_{0}$ & 0.3906 & 0.1058 \\
\hline
\end{tabularx}
\vspace{-2mm}
\end{table}

\begin{table}[!tbp]
\captionsetup{skip=2.5pt}
\caption{Ablation study on serialization scheme under the default \textit{singularity-aware} traversal (Circular relative offset vs.\ Block-offset).}
\label{tab:ablation_serialization}
\footnotesize
\setlength{\tabcolsep}{1.2mm}
\renewcommand{\arraystretch}{1.2}
\centering
\begin{tabular}{l|c|c}
\hline
Metrics & Block-offset & Circ. relative offset \\
\hline
Scaled Jacobian $\uparrow^1_{-1}$          & 0.7442 & 0.9829 \\
Watertightness $\uparrow^1_{0}$            & 0.6635 & 0.9996 \\
Manifoldness $\uparrow^1_{0}$              & 0.9993 & 0.9999 \\
Chamfer distance $\downarrow^{\infty}_{0}$  & 0.3161 & 0.0684 \\
Hausdorff distance $\downarrow^{\infty}_{0}$ & 0.5487 & 0.1058 \\
\hline
\end{tabular}
\vspace{-2mm}
\end{table}

\subsection{Ablation Study}\label{sec:Ablation}

\begin{figure}[!tbp]
    \centering
  \includegraphics[width=0.98\linewidth]{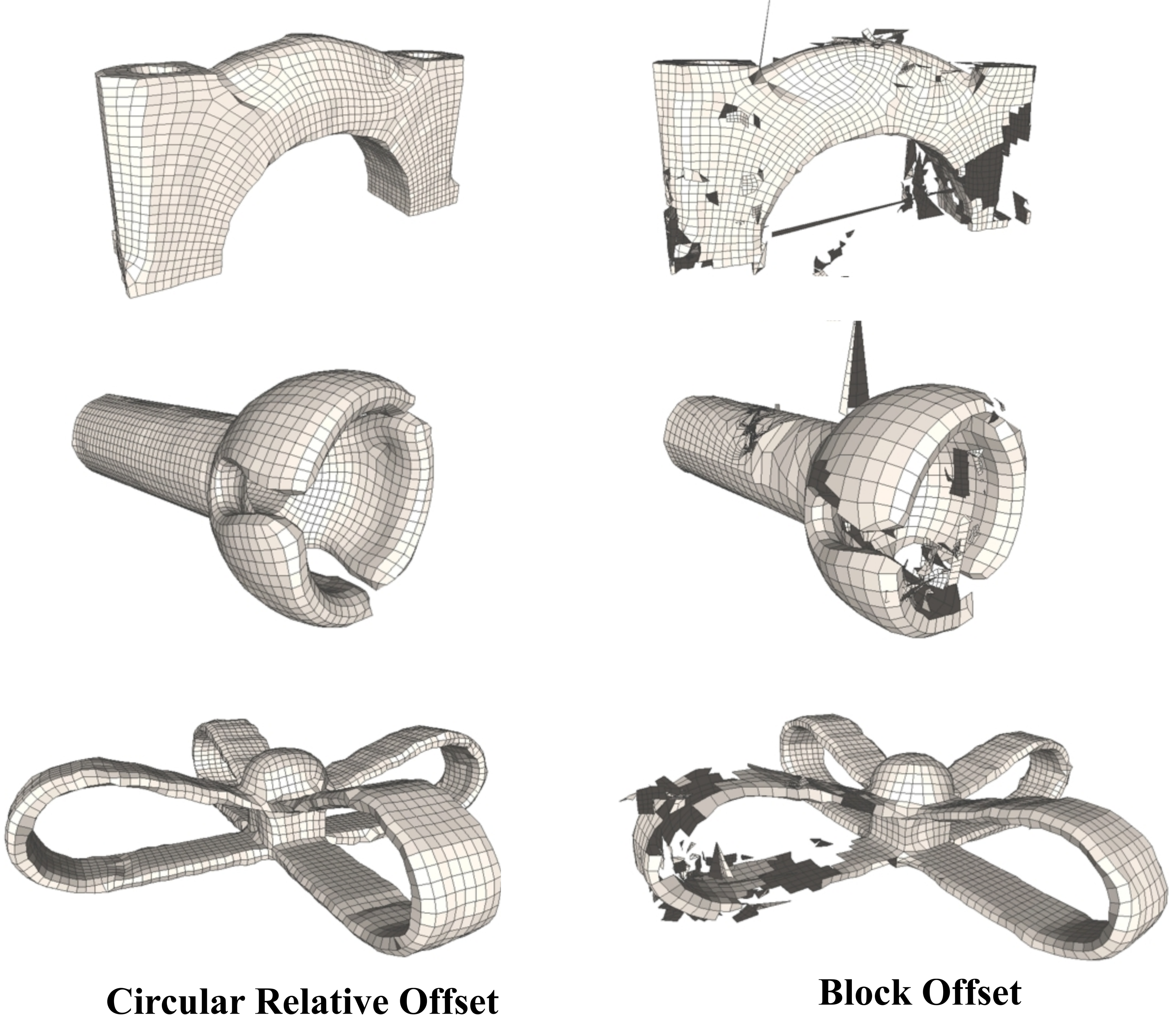}
    \caption{Qualitative ablation on serialization schemes.}
    \label{fig:serialization schemes}
    \vspace{-2mm}
\end{figure}

\begin{figure}[!tbp]
    \centering
    \includegraphics[width=0.98\linewidth]{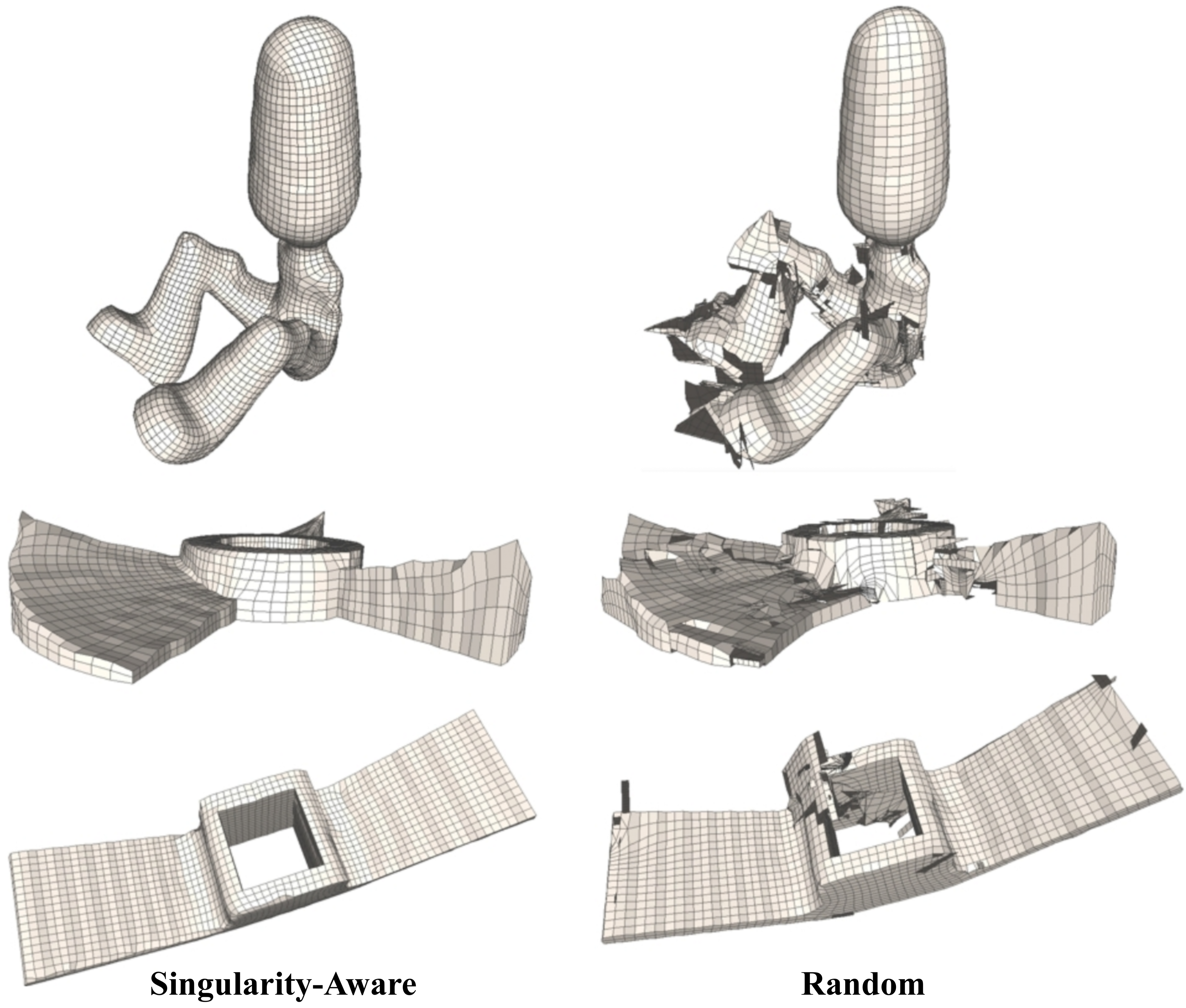}
    \caption{Qualitative ablation on patch traversal order.}
    \label{fig:traversal order}
    \vspace{-3mm}
\end{figure}

To validate the effectiveness of the core designs in our Q-MAGIC pipeline, we conduct ablation studies on the center vertex selection and the serialization scheme, with results reported in Table~\ref{tab:ablation_traversal} and Table~\ref{tab:ablation_serialization}. Qualitative visualizations are provided in Figure \ref{fig:serialization schemes} and Figure \ref{fig:traversal order}

Table~\ref{tab:ablation_traversal} shows that singularity-aware traversal consistently improves mesh validity and element quality over random traversal. Since singularity regions determine local valence distributions and edge-loop organization, errors introduced there are difficult to correct later. Random traversal often establishes connectivity in regular regions first, forcing the model to satisfy singularity constraints within an already generated structure, which can lead to disordered layouts and broken loops. In contrast, singularity-aware traversal anchors critical connections early, providing stronger structural guidance and reducing error accumulation.

Table~\ref{tab:ablation_serialization} shows that the circular relative offset representation is more robust than the block-offset representation. By shifting prediction from global positions to local displacements, circular relative offsets operate within a bounded value range and avoid the large jumps introduced by global discretization. In contrast, absolute offsets are more susceptible to drift across regions, and the resulting errors can be amplified during de-serialization, leading to reduced mesh validity and geometric accuracy.

We further analyze the computational sensitivity to the block size $B$ and offset size $O$. Increasing $(B,O)$ from the default $(8,16)$ to $(16,32)$ raises the per-axis spatial resolution from $128$ to $512$, while substantially enlarging the corresponding token spaces. Under the same training setting, the peak GPU memory increases and the training time increases by approximately \textbf{53\%}. Therefore, we adopt $(B,O)=(8,16)$ as a practical configuration that provides sufficient spatial resolution while maintaining manageable memory and training costs.

\section{Limitations and Future Work}
Despite the promising results, Q-MAGIC still has several limitations. As an autoregressive framework, its computational cost increases with sequence length, which may limit scalability to highly complex or high-resolution meshes. In addition, global connectivity across locally generated regions is not always guaranteed for complex shapes. Our current evaluation demonstrates good generalization to diverse shapes within the Thing10K distribution; however, cross-domain generalization to CAD-oriented or artist-designed quad meshes with different topology and edge-flow patterns remains to be further explored. Finally, the singularity-aware traversal currently relies primarily on vertex valence.

In future work, we plan to explore hierarchical decoding for better scalability, explicit global connectivity constraints for improved topological consistency, more diverse cross-domain training data for stronger generalization, and geometry-aware singularity selection incorporating curvature, feature lines, and boundary information.

\section{Conclusion}\label{sec13}

 In this paper, we introduce Q-MAGIC, an autoregressive Transformer framework for quad-only mesh generation, which takes the undifferentiable connectivity prediction as the next face token prediction task. To shorten token sequences and enhance the overall regularity of the global mesh, we introduce a singularity-aware serialization to prioritize topology-critical wiring and improve quad mesh validity. Furthermore, to compactly encode geometry and handle mesh-density variation, we propose circular relative offset tokenization, which encodes geometry via local relative offsets into discrete tokens, achieving mesh-density adaptation. Experimental results demonstrate that our Q-MAGIC outperforms baseline methods in both mesh validity and reconstruction quality.

\section*{Acknowledgment}
This research was supported by the National Key R\&D Program of China under Grant No. 2021YFA1003003, the Natural Science Foundation of China under Grant No. T2225012, 12494554, 12494550.



\bibliographystyle{eg-alpha-doi} 
\bibliography{egbibsample}       




\end{document}